\documentclass{article}
\usepackage{spconf,amsmath,epsfig,algorithm,algorithmic,multirow,slashbox}

\begin{document}\sloppy

\def\x{{\mathbf x}}
\def\L{{\cal L}}

\title{Learning deep representation from coarse to fine for face alignment}
%
\name{Zhiwen Shao, Shouhong Ding, Yiru  Zhao, Qinchuan  Zhang, and Lizhuang Ma\thanks{This work was sponsored by the National Natural Science Foundation of China (No. 61133009 and 61472245).}}
\address{Department of Computer Science and Engineering, Shanghai Jiao Tong University, China\\
\{shaozhiwen, feiben, yiru.zhao, qinchuan.zhang\}@sjtu.edu.cn, ma-lz@cs.sjtu.edu.cn}
%
%
%

\maketitle

\begin{abstract}
In this paper, we propose a novel face alignment method that trains deep convolutional network from coarse to fine. It divides given landmarks into principal subset and elaborate subset. We firstly keep a large weight for principal subset to make our network primarily predict their locations while slightly take elaborate subset into account. Next the weight of principal subset is gradually decreased until two subsets have equivalent weights. This process contributes to learn a good initial model and search the optimal model smoothly to avoid missing fairly good intermediate models in subsequent procedures. On the challenging COFW dataset \cite{burgos2013robust}, our method achieves $6.33\%$ mean error with a reduction of $21.37\%$ compared with the best previous result \cite{zhang2015learning}.
\end{abstract}
\section{Introduction}
\label{sec:intro}

Face alignment aims to locate facial landmarks such as eyes and noses automatically. It is a preprocessing stage for many facial analysis tasks like face verification \cite{sun2014deeply} and facial attributes analysis \cite{datta2011hierarchical}. Though great success has been achieved in this field recently, robust facial landmark detection remains a challenging problem in the presence of severe occlusion and large pose variations. Most conventional methods \cite{burgos2013robust,cao2014face,xiong2013supervised,ren2014face} are based on low-level features and have limited capacity to represent highly complex faces. Thus we use deep convolutional network which is effective in extracting features and robust to occlusions \cite{sun2014deeply}.

We discover that there are a few key landmarks which can coarsely determine face shape including brow corners, eye corners, nose tip, mouth corners and chin tip. We call a subset consists of these landmarks as principal subset, and the elaborate subset is made up of remaining landmarks. In the first step, we set a very large weight for principal subset and so the weight of elaborate subset is tiny. In this way, our network mainly predicts the location of principal subset while locates the elaborate subset roughly. During subsequent procedures, we gradually decrease the weight of principal subset, which helps to search the optimal model steadily without missing fairly good models. And entire landmarks are accurately located with the finally learned model.

Inspired by \cite{sun2014deeply}, we enhance the supervision by adding supervisory signal to each of the four max-pooling layers rather than only supervising the last max-pooling layer. And we employ an effective data augmentation strategy to overcome the lack of training images.

The remainder of this paper is organized as follows. In the next section, we discuss the related works of face alignment and analyse their characteristics. In Section \ref{sec:appro} , we elaborate our coarse-to-fine training algorithm (CFT) and illuminate the structure of our deep convolutional network, following which the implementation details is exhibited. Several comparative experiments are carried out in Section \ref{sec:experi} to show the precision and robustness of our model. Section \ref{sec:conclu} is the conclusion of this paper.

\section{Related work}
\label{sec:related}

Significant progress on face alignment has been achieved in recent years, including conventional methods and deep learning methods.

\textbf{Conventional methods:} Active appearance models (AAM) \cite{cootes2001active} reconstructs entire face using an appearance model and minimize the texture residual to estimate the shape. Supervised descent method (SDM) \cite{xiong2013supervised} aims at solving nonlinear least squares optimization problem, which applies non-linear SIFT \cite{lowe2004distinctive} feature and linear regressors. Both Cao et al. \cite{cao2014face} and Burgos-Artizzu et al. \cite{burgos2013robust} use boosted ferns to regress the shape increment with pixel-difference features.

These methods mainly refine the prediction of the landmarks location iteratively from an initial estimate, which is highly relevant to the initialization. In contrast, our network takes raw faces as input without any initialization.

\textbf{Deep learning methods:} Sun et al. \cite{sun2013deep} estimates the positions of facial landmarks with three-level cascaded convolutional networks. Zhang et al. \cite{zhang2014coarse} uses successive auto-encoder networks for face alignment. Both methods use multiple deep networks to locate the landmarks in a coarse-to-fine manner. They search the optimal location of landmarks from coarse to fine for each image. On the contrary, our method contains only one network and uses coarse-to-fine strategy during training.

Zhang et al. \cite{zhang2015learning} trains a deep convolutional network with multitask learning which jointly optimizes landmark detection together with the recognition of some facial attributes. It pre-trains the network by five landmarks and then fine-tunes to predict the dense landmarks. However, our method doesn't require labeling extra attributes for training samples. Different from pre-training, we also consider predicting the location of other elaborate landmarks. Compared to the method consists of pre-training and fine-tuning, we gradually adjust the weight of principal subset and elaborate subset respectively to avoid missing good models in subsequent training procedures.

\section{Our approach}
\label{sec:appro}

We propose a novel coarse-to-fine training algorithm with a good initialization and search optimal model smoothly. And our deep convolutional network has a strong ability to extract face features and predict landmarks location precisely.

\subsection{Coarse-to-fine training algorithm}
\label{ssec:algorithm}

\begin{figure}[!htb]
\centering\includegraphics[width=8cm]{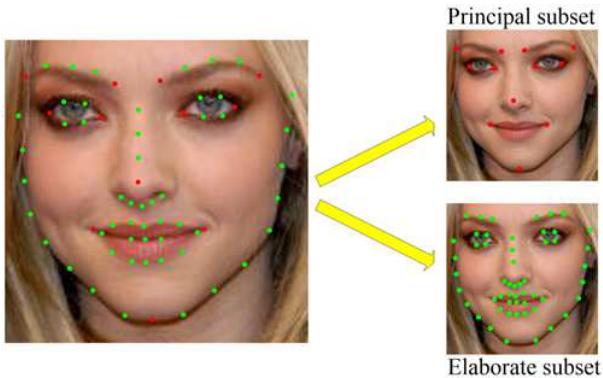}
\caption{Facial landmarks are divided into principal subset and elaborate subset. And the principal subset is made up of brow corners, eye corners, nose tip, mouth corners and chin tip.}
\label{fig:backbone_remainder}
\end{figure}

Since dense landmarks are expensive to label, directly trained model is apt to overfit to small training set. Therefore, we propose an innovative coarse-to-fine training algorithm. As shown in Figure \ref{fig:backbone_remainder}, given landmarks can be split into principal subset and elaborate subset. The former consists of twelve key points like eye corners, nose tip and mouth corners. Indeed pupils are also key points, but we don't choose them because many face alignment datasets such as Helen \cite{le2012interactive} and 300-W \cite{sagonas2013300} don't annotate pupils.

Our network directly outputs the coordinate of each landmark, so we need to minimize following loss function as
\begin{equation}
E=\lambda E_b + (1-\lambda) E_r,
\label{eq:loss}
\end{equation}
where $E_b$, $E_r$ refer to the loss of principal subset and elaborate subset respectively, which will be elaborated in Section \ref{ssec:network}. And $\lambda$ controls the relative weight of principal subset. Our training algorithm CFT is sketched in Algorithm \ref{alg:train_alg}. The trainable parameters $\Theta$ are the link weights between different layers in the network.

\begin{algorithm}[!htb]
\caption{Coarse-to-fine training algorithm.}
\label{alg:train_alg}
\begin{algorithmic}[1]
\REQUIRE Network $N$ with trainable initialized parameters $\Theta$, initial control parameter $\lambda_0$, stage number $k\geq2$.
\ENSURE Trainable parameters $\Theta$.
\FOR{$i=0$ to $k-1$}
    \STATE $\lambda=\lambda_0 - (\lambda_0 -0.5)/(k-1) \cdot i$;
    \WHILE{$\mbox{not convergence}$}
    \STATE Training $N$ with back propagation (BP) \cite{rumelhart1985learning} algorithm and update $\Theta$;
    \ENDWHILE
\ENDFOR
\end{algorithmic}
\end{algorithm}

In the beginning $\lambda$ is initialized with $\lambda_0$ which is nearly equal to $1$ but smaller than $1$. Thus, our network mainly predicts the location of principal landmarks while remaining elaborate landmarks are still considered with the weight $1-\lambda$. This initial step lays emphasis on key points, which is beneficial for extracting essential face features. With the reduction of $\lambda$, our network searches the optimal model parameters $\Theta$ steadily until $\lambda$ is equal to $0.5$.

\begin{figure}[!htb]
\centering\includegraphics[width=8cm]{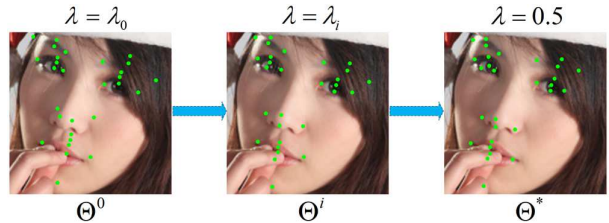}
\caption{Brief overview of our coarse-to-fine training algorithm. $\lambda$ and $\Theta$ beside each image denote the control parameter and corresponding learned model parameters respectively in each stage. $\Theta^*$ is finally learned model parameters.}
\label{fig:overview_alg}
\end{figure}

Figure \ref{fig:overview_alg} shows the overview of our training algorithm briefly. When finishing the first procedure, our network has been able to locate landmarks coarsely. It's clearly that the trained model is optimized stage by stage and the prediction of landmarks location using finally learned model $\Theta^*$ is very accurate.

\subsection{Deep convolutional network}
\label{ssec:network}

\begin{figure*}[!htb]
\centering\includegraphics[width=17cm]{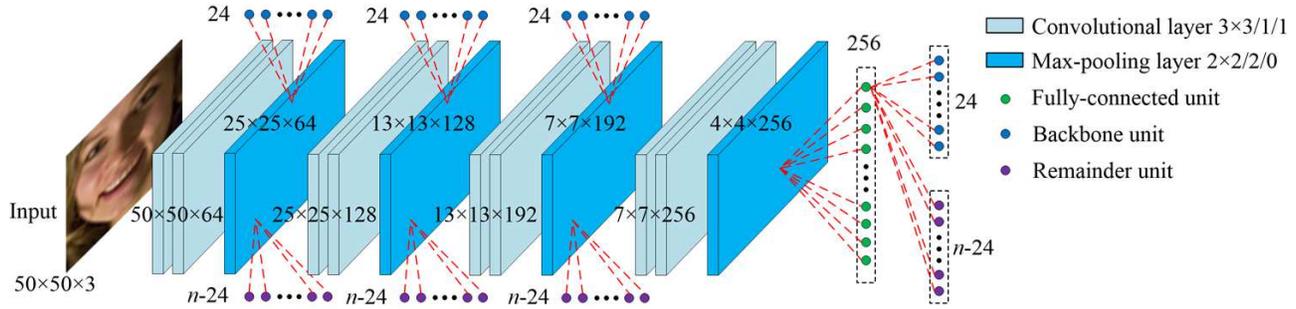}
\caption{The structure of our network. The equation $h \times w \times c$ beside each layer denotes that the size of map is $h \times w$ and the number of map is c. Every two continuous convolutional layers share the same equation. The equation $k_h \times k_w / s / p$ denotes that the filter size is $k_h \times k_w$, and the size of stride and padding for the filter are $s$ and $p$ respectively. Each convolutional layer has same filter parameters, equally applying to each max-pooling layer.}
\label{fig:netstruc}
\end{figure*}

Our deep convolutional network mainly comprises eight convolutional layers followed by one fully-connected layer and two split fully-connected layers on behalf of principal subset and elaborate subset respectively. And every two continuous convolutional layers connect with a max-pooling layer. Figure \ref{fig:netstruc} shows the detailed structure of our network which predicts the coordinate of each landmark in the final two split fully-connected layers. The input of our network is $50 \times 50 \times 3$ for color face patches. $n$ is equal to double total number of landmarks, e.g. $68\times2=136$ for 300-W dataset. Besides the size of corresponding fully-connected layer for principal subset is $12\times2=24$.

Suggested by \cite{sun2014deeply}, we enhance the supervision by adding supervisory signal to each of the four max-pooling layers rather than only supervising the last max-pooling layer. It should be noted that the first three split fully-connected layers are connected with corresponding max-pooling layers directly without using a fully-connected layer as intermediate. The reason is that we only need to add supervisory signal to former layers, while regard final output as the prediction of landmarks location. The penultimate layer with $256$ units is conducive to extract global high-level features for more accurate prediction from the final two split fully-connected layers.

In order to accelerate the training of network, we add a batch normalization layer \cite{ioffe2015batch} after each convolutional layer. Batch normalization is scaling and shifting the normalized input as
\begin{equation}
y=\gamma \hat{x}+\beta,
\label{eq:batchnormaliz}
\end{equation}
where $\hat{x}=\frac{x-E[x]}{\sqrt{Var[x]}}$, the expectation and variance are computed over a mini-batch from the training dataset. After normalizing each convolutional layer, ReLU nonlinearity ($y=max(0,x)$) is added to speed up convergence. We don't operate ReLU on the penultimate fully-connected layer and every two split fully-connected layers in order to preserve important information. It is worth mentioning that our network is based on VGG net \cite{simonyan2014very} whose stacked multiple convolutional layers jointly form complex features.

All the four supervisory signals use same loss function defined in Equation \ref{eq:loss}. Our approach is evaluated based on alignment error measured with the distance between estimated coordinate and ground truth coordinate normalized by the inter-ocular distance. So we use normalized Euclidean distance rather than straightforward Euclidean distance to calculate loss, which is formulated as
\begin{equation}
E_b=\frac{\Vert f_b-\hat{f_b} \Vert_2^2}{2d^2},
\label{eq:eucli1}
\end{equation}
\begin{equation}
E_r=\frac{\Vert f_r-\hat{f_r} \Vert_2^2}{2d^2},
\label{eq:eucli2}
\end{equation}
where $f_b$ and $f_r$ are the vector concatenating the ground truth coordinate of all landmarks belonging to principal subset and elaborate subset respectively, and $\hat{f_b}$ and $\hat{f_r}$ denote predicted landmarks location correspondingly. $d$ is the inter-ocular distance. And the coefficient $\frac{1}{2}$ makes the derivation of loss more convenient during back propagation.
\subsection{Implementation details}
\label{ssec:imple_detail}

Before starting the face alignment, we need to carry out face detection on the training images as preprocessing. Then we can acquire a face bounding box which is used for taking face patches. We conduct data augmentation since benchmark face alignment datasets such as Helen, 300-W and COFW have too small training sets.

We rotate the face image with different angles and determine new face bounding box, then slightly translate the face bounding box ensured to contain all the landmarks. It is worth mentioning that the new area contained in the face bounding box is derived from the original image rather than artificial setting which may has bad impacts on the training process.

The translation operation helps to improve the robustness of landmark detection in the condition of tiny face shift, especially in face tracking. And our model can learn to adapt complex pose variation thanks to the rotation operation. In the next steps, we horizontally flip each face patch and finally conduct JPEG compression. So our network will be trained to be robust to poor-quality images which is ubiquitous in the real case.

\begin{table*}[!htb]
\centering\caption{Comparison of mean errors (\%) with state-of-the-art methods. DT is training our network based on conventional direct training algorithm. The results of state-of-the-art methods are obtained directly from the literatures. It's worth to mention that results of some earlier methods are provided by recent papers.}
\label{tab:comp_other_tab}
\begin{tabular}{|*{7}{c|}}
\hline
\multirow{2}*{Method}&\multicolumn{2}{|c|}{Helen}&\multicolumn{3}{|c|}{300-W}&\multicolumn{1}{|c|}{COFW}\\
\cline{2-7}&194 landmarks&68 landmarks&Common Subset&Challenging Subset&Fullset&29 landmarks\\
\hline
ESR \cite{cao2014face} &5.70 &- &5.28 &17.00 &7.58 &11.2 \\
RCPR \cite{burgos2013robust} &6.50 &5.93 &6.18 &17.26 &8.35 &8.5 \\
SDM \cite{xiong2013supervised} &5.85 &5.50 &5.57 &15.40 &7.50 &11.14 \\
LBF \cite{ren2014face} &5.41 &- &4.95 &11.98 &6.32 &- \\
CFAN \cite{zhang2014coarse} &- &5.53 &5.50 &16.78 &7.69 &-\\
ERT \cite{kazemi2014one} &4.90 &- &- &- &6.40 &- \\
CFSS \cite{zhu2015face} &4.74 &4.63 &4.73 & 9.98 &5.76 &- \\
TCDCN \cite{zhang2015learning} &4.63 &4.60 &4.80 &8.60 &5.54 &8.05 \\
\textbf{DT} &5.32 &5.21 &5.25 &10.42 &6.26 &6.75\\
\textbf{CFT} &4.86 &4.75 &4.82 &10.06 &5.85 &\textbf{6.33}\\
\hline
\end{tabular}
\end{table*}

In this way, the training face patches are increased by many times. We train our network using a deep learning framework Caffe \cite{jia2014caffe}, and control parameter $\lambda_0$ and $k$ are set to be $0.995$ and $3$ respectively. If stage number $k$ is too large, training our network will be time-consuming. So assigning $3$ to $k$ balances the time and accuracy suitably.

\section{Experiments}
\label{sec:experi}

We firstly investigate the advantages and effectiveness of our coarse-to-fine training algorithm by comparing to ordinary algorithm. Then we compare our method CFT against state-of-the-art methods on three widely used benchmark datasets, Helen, 300-W and COFW. For each dataset we report the inter-ocular distance normalized error averaged over all landmarks and images, similar to most previous works.

\textbf{Helen} contains $2000$ training images and $330$ testing images, annotated densely with $194$ landmarks. We also evaluate on $68$ landmarks provided by \cite{sagonas2013300}.

\textbf{300-W} is created from existing datasets, including AFW \cite{zhu2012face}, LFPW \cite{belhumeur2013localizing}, Helen and XM2VTS \cite{messer1999xm2vtsdb} with annotation of $68$ landmarks. In addition, it contains a challenging IBUG set. As in \cite{ren2014face}, our training set consists of AFW, the training sets of LFPW and Helen with $3148$ images totally. And we perform testing with three forms: the test images from LFPW and Helen as the common subset, the IBUG as the challenging subset, and the union of them as the full set with $689$ images in all.

\textbf{COFW} contains $1007$ images annotated with $29$ landmarks collected from the web. It is designed to present faces with large variations in shape and occlusions due to differences in pose, expression, use of accessories and interactions with objects. The training set consists of $845$ LFPW faces and $500$ COFW faces ($1345$ total), and the testing set contains remaining $507$ COFW faces.

\subsection{Algorithm discussions}
\label{ssec:alg_discuss}

The conventional training algorithm for deep convolutional network learns overall features directly, namely $\lambda=0.5$ with only one procedure, based on randomly initialized parameters derived from a distribution. We call the conventional algorithm as direct training algorithm (DT). In contrast, our training algorithm CFT firstly concentrates on learning features of principal subset based on randomly initialized parameters. In this paper, we use standard Gaussian distribution to randomly initialize model parameters. Next, we gradually decrease the weight of principal subset and increase the weight of elaborate subset simultaneously until both are equivalent. In this way, our network can search models from coarse to fine smoothly.

\begin{figure}[!htb]
\centering\includegraphics[width=8cm]{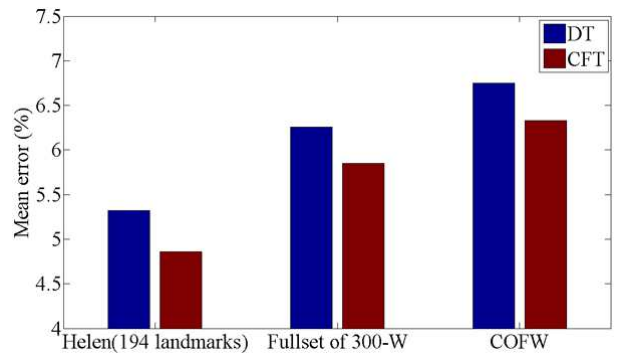}
\caption{Comparison between CFT and DT tested on Helen($194$ landmarks), Fullset of 300-W and COFW respectively.}
\label{fig:conven_ours}
\end{figure}

\begin{figure*}[!htb]
\centering\includegraphics[width=17cm]{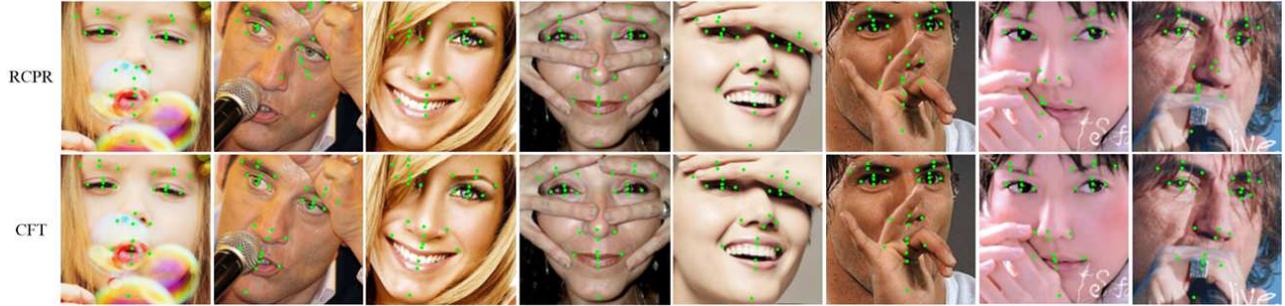}
\caption{The results of RCPR and CFT on several challenging images from COFW.}
\label{fig:cofw}
\end{figure*}

In order to compare CFT and DT, we also train our network using DT and test on benchmark alignment datasets. The comparison between CFT and DT is shown in Figure \ref{fig:conven_ours} and Table \ref{tab:comp_other_tab}. Our algorithm CFT outperforms DT by a large margin, with a reduction of $8.65\%$, $6.55\%$, $6.22\%$ on Helen($194$ landmarks), Fullset of 300-W and COFW respectively.

\subsection{Comparison with other methods}
\label{ssec:comp_other}

We evaluate our approach CFT on Helen, 300-W and COFW based on mean error normalised by the inter-ocular distance. And we compare with state-of-the-art methods including ESR \cite{cao2014face}, RCPR \cite{burgos2013robust}, SDM \cite{xiong2013supervised}, LBF \cite{ren2014face}, CFAN \cite{zhang2014coarse}, ERT \cite{kazemi2014one}, CFSS \cite{zhu2015face} and TCDCN \cite{zhang2015learning} as shown in Table \ref{tab:comp_other_tab}. In particular, TCDCN uses outside training images during the pre-training stage. This is unfair for other methods including ours which only use the training images provided by benchmark face alignment datasets.

It is obvious that CFT outperforms most of the state-of-the-art methods. Although the mean error of CFT tested on Helen and 300-W is slightly higher than CFSS and TCDCN, CFT performs better on challenging COFW whose faces are taken with severe occlusion. Specifically, CFT significantly reduces the error by $21.37\%$ on the challenging COFW in comparison to the state-of-the-art TCDCN.

Sun et al. \cite{sun2014deeply} prove that deep convolutional network is robust to occlusions, thus our approach can better take advantage of the superiority of deep convolutional network than TCDCN which is also based on deep convolutional network. TCDCN trains a deep convolutional network with multitask learning which jointly optimizes landmark detection together with the recognition of some facial attributes. It pre-trains the network by five landmarks and then fine-tunes to predict the dense landmarks.

Nevertheless, our method doesn't require labeling extra attributes for training samples. Different from pre-training, we also consider predicting the location of other landmarks when laying emphasis on the principal subset. Compared to the method consists of pre-training and fine-tuning, we gradually adjust the weight of principal subset and elaborate subset respectively to avoid missing optimal models in subsequent procedures.

\begin{figure}[!htb]
\centering\includegraphics[width=8cm]{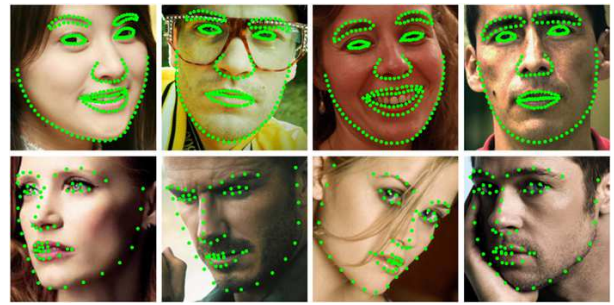}
\caption{Example alignment results on Helen ($194$ landmarks) and IBUG.}
\label{fig:helen_ibug}
\end{figure}

Figure \ref{fig:cofw} shows several examples of landmark detection using RCPR and CFT respectively. It's clearly that our approach exhibits superior capability of handling complex occlusion, thanks to the good model trained by coarse-to-fine training algorithm. We also provide examples of alignment results of our approach on Helen and challenging subset IBUG of 300-W in Figure \ref{fig:helen_ibug}. It can be observed that our approach can locate landmarks accurately in the condition of complex pose, illumination and expression variations.

Our method takes $65$ ms to process an image on a single Intel Core i7-4790 CPU. This speed is slower than 18ms of TCDCN because our network is more complicated. We will try to reduce the complexity of our network in further research.

\section{Conclusion}
\label{sec:conclu}

We propose a novel coarse-to-fine training algorithm to train deep convolutional network for facial landmark detection. This algorithm contributes to search the optimal model steadily without missing fairly good models. Our network directly predicts the coordinates of landmarks using single network without any other additional operations, whilst significantly improves the accuracy of face alignment in the condition of severe occlusion. And we believe that the proposed training algorithm can also be applied to other problems with the use of deep convolutional network.

\bibliographystyle{IEEEbib}
\bibliography{icme2016}

\end{document}